\documentclass[runningheads]{llncs}

\usepackage{color}
\usepackage{graphicx}
\usepackage{amsmath}
\usepackage{amssymb}
\usepackage[labelformat=simple]{subfig}
\usepackage[ruled,lined,linesnumbered]{algorithm2e}
\usepackage{algorithmic}
\usepackage{blindtext}
\usepackage{enumitem}
\usepackage{multirow}
\usepackage[labelformat=simple]{subfig}
\usepackage{threeparttable/threeparttable}
\usepackage{bm}
\usepackage[normalem]{ulem}
\usepackage{hyperref}
\usepackage{amssymb}
\usepackage{pifont}

\newcommand{\etal}{\mbox{\emph{et al.}}}
\newcommand{\ie}{\mbox{\emph{i.e.,\ }}}
\newcommand{\bd}[1]{\textbf{#1}}

\newcommand{\tpm}[1]{\resizebox{4mm}{!}{\Large\raisebox{1mm}{$\genfrac{}{}{0pt}{}{}{\pm#1}$}}}

\usepackage{array}
\usepackage{hyperref}

\newcolumntype{P}[1]{>{\centering\arraybackslash}p{#1}}

\begin{document}
\title{NablaNet: A Nested Neural Network Architecture\\for Medical Image Segmentation}
\title{$\nabla$-Net: A Nested Ensemble of Neural Networks for Medical Image Segmentation}
\title{$\nabla$-Net: A Upside-down Triangulated Neural Network Architecture for Medical Image Segmentation}
\title{UNet++: A Nested U-Net Architecture\\for Medical Image Segmentation}
%
\titlerunning{UNet++: A Nested U-Net Architecture}
%
\author{
    Zongwei Zhou \and
    Md Mahfuzur Rahman Siddiquee \and \\
    Nima Tajbakhsh \and
    Jianming Liang
}
\authorrunning{Z. Zhou, \etal}

\institute{
    Arizona State University\\
    \email{\{zongweiz,mrahmans,ntajbakh,jianming.liang\}@asu.edu}
}
\maketitle              

\begin{abstract}
  In this paper, we present UNet++, a new, more powerful architecture for medical image segmentation. Our architecture is essentially a deeply-supervised encoder-decoder network where the encoder and decoder sub-networks are connected through a series of nested, dense  skip pathways. The re-designed skip pathways aim at reducing the semantic gap between the feature maps of the encoder and decoder sub-networks. We argue that the optimizer would deal with an easier learning task when the feature maps from the decoder and encoder networks are semantically similar. We have evaluated UNet++ in comparison with U-Net and wide U-Net architectures across multiple medical image segmentation tasks: nodule segmentation in the low-dose CT scans of chest, nuclei segmentation in the microscopy images, liver segmentation in abdominal CT scans, and polyp segmentation in colonoscopy videos. Our experiments demonstrate that UNet++ with deep supervision achieves an average IoU gain of 3.9 and 3.4 points over U-Net and wide U-Net, respectively.

\end{abstract}

\section{Introduction}
\label{sec:introduction}

The state-of-the-art models for image segmentation are variants of the encoder-decoder architecture like U-Net~\cite{ronneberger2015u} and fully convolutional network (FCN)~\cite{long2015fully}. These encoder-decoder networks used for segmentation share a key similarity: skip connections, which combine deep, semantic, coarse-grained feature maps from the decoder sub-network with shallow, low-level, fine-grained feature maps from the encoder sub-network. The skip connections have proved effective in recovering fine-grained details of the target objects; generating segmentation masks with fine details even on complex background. Skip connections is also fundamental to the success of instance-level segmentation models such as Mask-RCNN, which enables the segmentation of occluded objects. Arguably, image segmentation in natural images has reached a satisfactory level of performance, but do these models meet the strict segmentation requirements of medical images?

Segmenting lesions or abnormalities in medical images demands a higher level of accuracy than what is desired in natural images. While a precise segmentation mask may not be critical in natural images, even marginal segmentation errors in medical images can lead to poor user experience in clinical settings. For instance, the subtle spiculation patterns around a nodule may indicate nodule malignancy; and therefore, their exclusion from the segmentation masks would lower the credibility of the model from the clinical perspective. Furthermore, inaccurate segmentation may also lead to a major change in the subsequent computer-generated diagnosis. For example, an erroneous measurement of nodule growth in longitudinal studies can result in the assignment of an incorrect Lung-RADS category to a screening patient. It is therefore desired to devise more effective image segmentation architectures that can effectively recover the fine details of the target objects in medical images.

To address the need for more accurate segmentation in medical images, we present UNet++, a new segmentation architecture based on nested and dense skip connections.
The underlying hypothesis behind our architecture is that the model can more effectively capture fine-grained details of the foreground objects when high-resolution feature maps from the encoder network are gradually enriched prior to fusion with the corresponding semantically rich feature maps from the decoder network. We argue that the network would deal with an easier learning task when the feature maps from the decoder and encoder networks are semantically similar. This is in contrast to the plain skip connections commonly used in U-Net, which directly fast-forward high-resolution feature maps from the encoder to the decoder network, resulting in the fusion of semantically dissimilar feature maps. According to our experiments, the suggested architecture is effective, yielding significant performance gain over U-Net and wide U-Net.

\section{Related Work}
\label{sec:related_works}

Long~\etal~\cite{long2015fully} first introduced fully convolutional networks (FCN), while U-Net was introduced by Ronneberger~\etal~\cite{ronneberger2015u}. They both share a key idea: skip connections.  In FCN, up-sampled feature maps are summed with feature maps skipped from the encoder, while U-Net concatenates them and add convolutions and non-linearities between each up-sampling step. The skip connections have shown to help recover the full spatial resolution at the network output, making fully convolutional methods suitable for semantic segmentation. Inspired by DenseNet architecture~\cite{huang2017densely}, Li~\etal~\cite{li2017h} proposed H-denseunet for liver and liver tumor segmentation. In the same spirit, Drozdzal\etal~\cite{drozdzal2016importance} systematically investigated the importance of skip connections, and introduced short skip connections within the encoder. Despite the minor differences between the above architectures, they all tend to fuse semantically dissimilar feature maps from the encoder and decoder sub-networks, which, according to our experiments, can degrade segmentation performance.

The other two recent related works are GridNet~\cite{fourure2017residual} and Mask-RCNN~\cite{he2017mask}. GridNet is an encoder-decoder architecture wherein the feature maps are wired in a grid fashion, generalizing several classical segmentation architectures. GridNet, however, lacks up-sampling layers between skip connections; and thus, it does not represent UNet++. Mask-RCNN is perhaps the most important meta framework for object detection, classification and segmentation. We would like to note that UNet++ can be readily deployed as the backbone architecture in Mask-RCNN by simply replacing the plain skip connections with the suggested nested dense skip pathways. Due to limited space, we were not able to include results of Mask RCNN with  UNet++ as the backbone architecture; however, the interested readers can refer to the supplementary material for further details.

\section{Proposed Network Architecture: UNet++}
\label{sec:methods}

\begin{figure}[t]
\begin{center}
\includegraphics[width=1.0\linewidth]{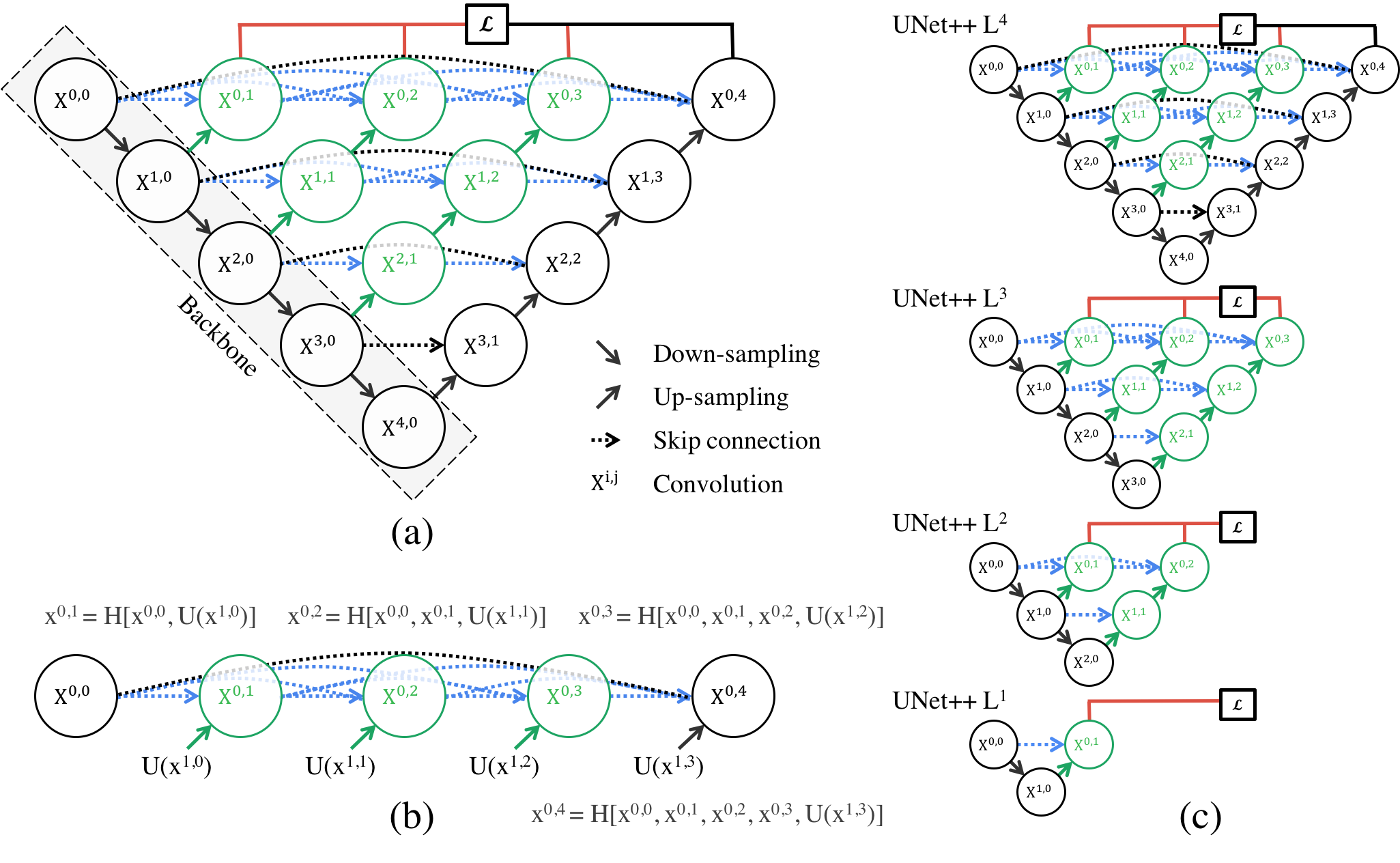}
\end{center}
\caption{(a) UNet++ consists of an encoder and decoder that are connected through a series of nested dense convolutional blocks. The main idea behind UNet++ is to bridge the semantic gap between the feature maps of the encoder and decoder prior to fusion. For example, the semantic gap between (X$^{0,0}$,X$^{1,3}$) is bridged using a dense convolution block with three convolution layers. In the graphical abstract, black indicates the original U-Net, green and blue show dense convolution blocks on the skip pathways, and red indicates deep supervision. Red, green, and blue components distinguish  UNet++ from U-Net. (b) Detailed analysis of the first skip pathway of UNet++. (c) UNet++ can be pruned at inference time, if trained with deep supervision. }
\label{fig:network_architecture}
\end{figure}

\figurename~\ref{fig:network_architecture}a shows a high-level overview of the suggested architecture. As seen, UNet++ starts with an encoder sub-network or backbone followed by a decoder sub-network. What distinguishes UNet++ from U-Net (the black components in \figurename~\ref{fig:network_architecture}a) is the re-designed skip pathways (shown in green and blue) that connect the two sub-networks and the use of deep supervision (shown red).

\subsection{Re-designed skip pathways}
Re-designed skip pathways transform the connectivity of the encoder and decoder sub-networks. In U-Net, the feature maps of the encoder are directly received in the decoder; however, in  UNet++, they undergo a dense convolution block whose number of convolution layers depends on the pyramid level. For example, the skip pathway between nodes X$^{0,0}$ and X$^{1,3}$ consists of a dense convolution block with three convolution layers where each convolution layer is preceded by a concatenation layer that fuses the output from the previous convolution layer of the same dense block with the corresponding up-sampled output of the lower dense block. Essentially, the dense convolution block brings the semantic level of the encoder feature maps closer to that of the feature maps awaiting in the decoder. The hypothesis is that the optimizer would face an easier optimization problem when the received encoder feature maps and the corresponding decoder feature maps are semantically similar.

Formally, we formulate the skip pathway as follows: let $x^{i,j}$ denote the output of node X$^{i,j}$ where $i$ indexes the down-sampling layer along the encoder and $j$ indexes the convolution layer of the dense block along the skip pathway. The stack of feature maps represented by $x^{i,j}$ is computed as

\begin{equation}
    \label{eq_unet}
    x^{i,j}=\begin{cases}
      \mathcal{H}\left(x^{i-1,j}\right),  & j=0  \\
      \mathcal{H}\left(\left[\left[x^{i,k}\right]_{k=0}^{j-1}, \mathcal{U}(x^{i+1,j-1}) \right]\right), & j>0  \\
    \end{cases}
\end{equation}

\noindent where function $\mathcal{H}(\cdot)$ is a convolution operation followed by an activation function, $\mathcal{U}(\cdot)$ denotes an up-sampling layer, and $[$ $]$ denotes the concatenation layer. Basically, nodes at level $j=0$  receive only one input from the previous layer of the encoder; nodes at level $j=1$ receive two inputs, both from the encoder sub-network but at two consecutive levels; and nodes at level $j>1$ receive $j+1$ inputs, of which $j$ inputs are the outputs of the previous $j$ nodes in the same skip pathway and the last input is the up-sampled output from the lower skip pathway. The reason that all prior feature maps accumulate and arrive at the current node is because we make use of a dense convolution block along each skip pathway. \figurename~\ref{fig:network_architecture}b further clarifies Eq.~\ref{eq_unet} by showing how the feature maps travel through the top skip pathway of UNet++.

\subsection{Deep supervision}
We propose to use deep supervision~\cite{lee2015deeply} in UNet++, enabling the model to operate in two modes: 1) accurate mode wherein the outputs from all segmentation branches are averaged; 2) fast mode wherein the final segmentation map is selected from only one of the segmentation branches, the choice of which determines the extent of model pruning and speed gain.  \figurename~\ref{fig:network_architecture}c shows how the choice of segmentation branch in fast mode results in architectures of varying complexity.

Owing to the nested skip pathways, UNet++ generates full resolution feature maps at multiple semantic levels, $\{x^{0,j},j\in\{1,2,3,4\}\}$, which are amenable to deep supervision. We have added a combination of binary cross-entropy and dice coefficient as the loss function to each of the above four semantic levels, which is described as:

\begin{equation}
\label{eq:object_function}
\mathcal{L}(Y,\hat{Y}) = -\frac{1}{N}\sum_{b=1}^{N}{\left(\frac{1}{2}\cdot Y_b\cdot\log{\hat{Y}_b}+\frac{2\cdot Y_b\cdot \hat{Y}_b}{Y_b+\hat{Y}_b}\right)}
\end{equation}

\noindent where $\hat{Y}_b$ and $Y_b$ denote the flatten predicted probabilities and the flatten ground truths of $b^{th}$ image respectively, and $N$ indicates the batch size.

In summary, as depicted in \figurename~\ref{fig:network_architecture}a, UNet++ differs from the original U-Net in three ways: 1) having convolution layers on skip pathways (shown in green), which bridges the semantic gap between encoder and decoder feature maps; 2) having dense skip connections on skip pathways (shown in blue), which improves gradient flow; and 3) having deep supervision (shown in red), which as will be shown in Section~\ref{sec:experiments} enables model pruning and improves or in the worst case achieves comparable performance to using only one loss layer.

\section{Experiments}
\label{sec:experiments}

\noindent{\bf{Datasets:}} As shown in Table~\ref{tab:dataset}, we use four medical imaging datasets for model evaluation, covering lesions/organs from different medical imaging modalities. For further details about datasets and the corresponding data pre-processing, we refer the readers to the supplementary material.

\begin{table}[t]
\centering
\caption{The image segmentation datasets used in our experiments.}
\label{tab:dataset} %
\begin{tabular}{P{0.16\linewidth}P{0.16\linewidth}P{0.18\linewidth}P{0.18\linewidth}P{0.28\linewidth}}
\hline
Dataset & Images & Input Size & Modality & Provider \\
\hline
cell nuclei & 670 & 96$\times$96 & microscopy & \tiny{\href{https://www.kaggle.com/c/data-science-bowl-2018}{Data Science Bowl 2018}} \\
colon polyp & 7,379 & 224$\times$224 & RGB video & \tiny{ASU-Mayo}~\cite{tajbakhsh2016convolutional,Zhou_2017_CVPR} \\
liver & 331  & 512$\times$512 & CT & \tiny{\href{https://competitions.codalab.org/competitions/17094}{MICCAI 2018 LiTS Challenge}} \\
lung nodule & 1,012 & 64$\times$64$\times$64 & CT & \tiny{\href{https://wiki.cancerimagingarchive.net/display/Public/LIDC-IDRI}{LIDC-IDRI}}~\cite{armato2011lung}\\
\hline
\end{tabular}
\end{table}


\vspace{4pt}
\noindent{\bf{Baseline models:}} For comparison, we used the original U-Net and a customized wide U-Net architecture. We chose U-Net because it is a common performance baseline for image segmentation. We also designed a wide U-Net with similar number of parameters as our suggested architecture. This was to ensure that the performance gain yielded by our architecture is not simply due to increased number of parameters. Table~\ref{tab:wide-unet} details the U-Net and wide U-Net architecture.


\begin{table}[t]
\centering
\caption{Number of convolutional kernels in U-Net and wide U-Net.}
\label{tab:wide-unet} %
\begin{tabular}{P{0.25\linewidth}P{0.14\linewidth}P{0.14\linewidth}P{0.14\linewidth}P{0.14\linewidth}P{0.14\linewidth}}
\hline
encoder / decoder & X$^{0,0}$/X$^{0,4}$ & X$^{1,0}$/X$^{1,3}$ & X$^{2,0}$/X$^{2,2}$ & X$^{3,0}$/X$^{3,1}$ & X$^{4,0}$/X$^{4,0}$ \\
\hline
U-Net & 32 & 64 & 128 & 256 & 512 \\
wide U-Net & 35 & 70 & 140 & 280 & 560 \\
\hline
\end{tabular}
\end{table}


\vspace{4pt}
\noindent{\bf{Implementation details:}} We monitored the Dice coefficient and Intersection over Union (IoU), and used {\em early-stop} mechanism on the validation set. We also used Adam optimizer with a learning rate of 3e-4. Architecture details for U-Net and wide U-Net are shown in Table~\ref{tab:wide-unet}. UNet++ is constructed from the original U-Net architecture.  All convolutional layers along a skip pathway (X$^{i,j}$) use $k$ kernels of size 3$\times$3 (or 3$\times$3$\times$3 for 3D lung nodule segmentation) where $k=32\times 2^i$. To enable deep supervision, a 1$\times$1 convolutional layer followed by a sigmoid activation function was appended to each of the target nodes: $\{x^{0,j}$ $|$ $j\in\{1,2,3,4\}\}$. As a result, UNet++ generates four segmentation maps given an input image, which will be further averaged to generate the final segmentation map. More details can be founded at \href{https://github.com/MrGiovanni/Nested-UNet}{github.com/Nested-UNet}.

\begin{figure}[t]
\begin{center}
\includegraphics[width=0.96\linewidth]{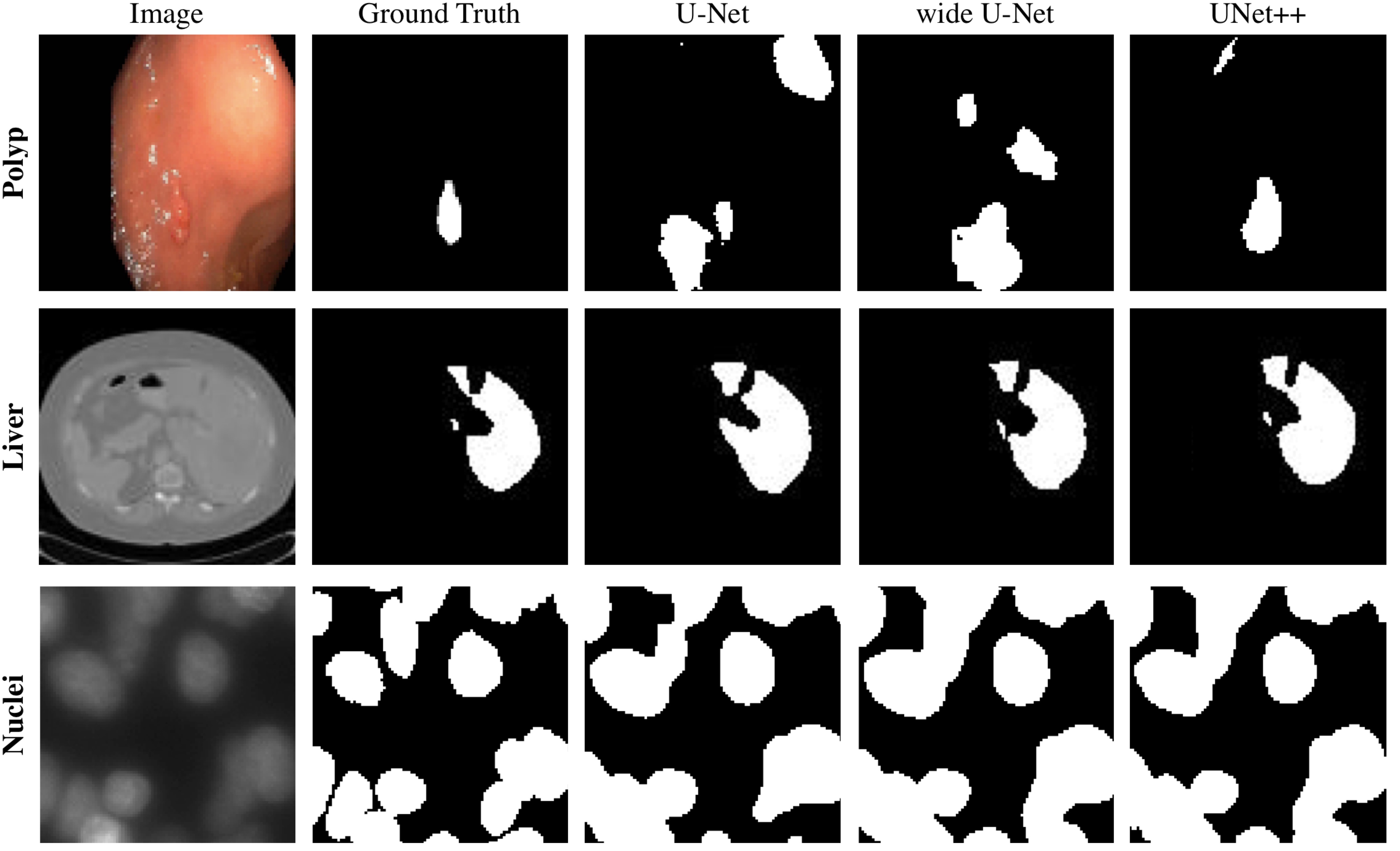}
\end{center}
\caption{Qualitative comparison between U-Net, wide U-Net, and UNet++, showing segmentation results for polyp, liver, and cell nuclei datasets (2D-only for a distinct visualization). }
\label{fig:predict_visualization}
\end{figure}

\vspace{4pt}
\noindent{\bf{Results:}}
Table~\ref{tab:main_results} compares U-Net, wide U-Net, and UNet++ in terms of the number parameters and segmentation accuracy for the tasks of lung nodule segmentation, colon polyp segmentation, liver segmentation, and cell nuclei segmentation. As seen, wide U-Net consistently outperforms U-Net except for liver segmentation where the two architectures perform comparably. This improvement is attributed to the larger number of parameters in wide U-Net. UNet++ without deep supervision achieves a significant performance gain over both U-Net and wide U-Net, yielding  average improvement of 2.8 and 3.3 points in IoU. UNet++ with deep supervision exhibits average improvement of 0.6 points over UNet++ without deep supervision. Specifically,  the use of deep supervision leads to marked improvement for liver and lung nodule segmentation, but such improvement vanishes for cell nuclei and colon polyp segmentation. This is because polyps and liver appear at varying scales in video frames and CT slices; and thus, a multi-scale approach using all segmentation branches (deep supervision) is essential for accurate segmentation. \figurename~\ref{fig:predict_visualization} shows a qualitative comparison between the results of U-Net, wide U-Net, and UNet++.

\begin{figure}[t]
\begin{center}
\includegraphics[width=1.0\linewidth]{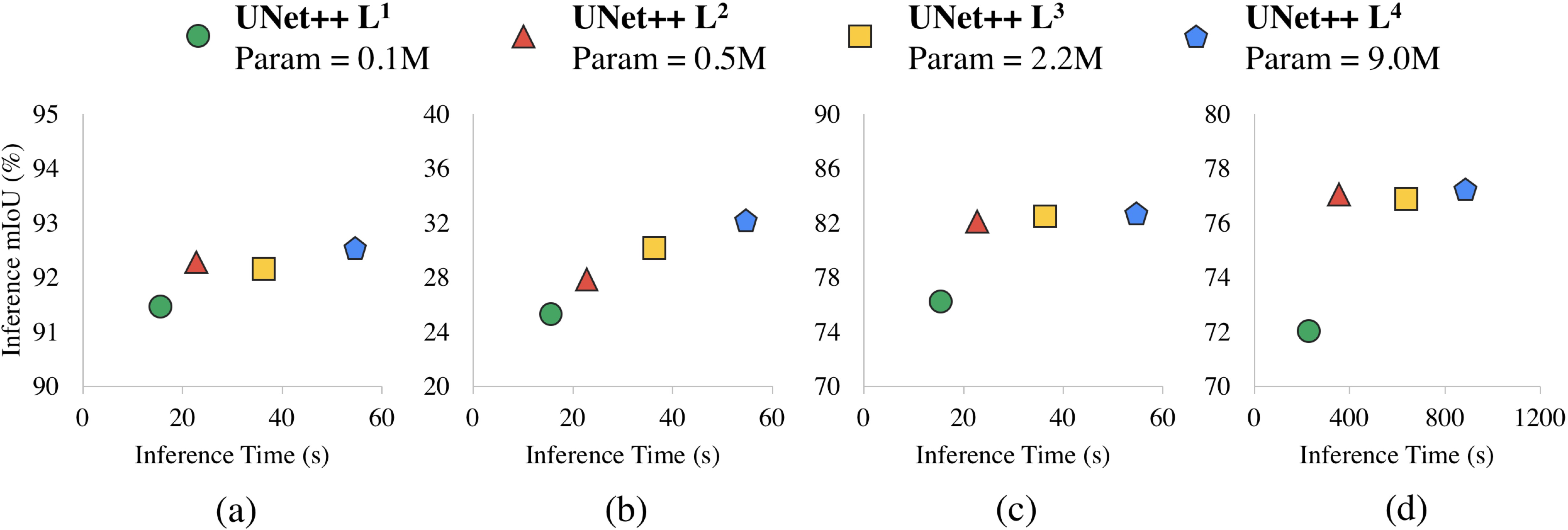}
\end{center}
\caption{Complexity, speed, and accuracy of UNet++ after pruning on (a) cell nuclei, (b) colon polyp, (c) liver, and (d) lung nodule segmentation tasks respectively. The inference time is the time taken to process {\bf 10k} test images using one NVIDIA TITAN X (Pascal) with 12 GB memory.}
\label{fig:prune_network}
\end{figure}

\vspace{4pt}
\noindent{\bf{Model pruning:}} Fig.~\ref{fig:prune_network} shows  segmentation performance of UNet++ after applying different levels of pruning. We use UNet++ L$^{i}$ to denote UNet++ pruned at level $i$ (see \figurename~\ref{fig:network_architecture}c for further details). As seen, UNet++ L$^{3}$ achieves on average 32.2\% reduction in inference time while degrading IoU by only 0.6 points.  More aggressive pruning further reduces the inference time but at the cost of significant accuracy degradation.

\begin{table}[t]
\begin{center}
\begin{threeparttable}
\caption{Segmentation results (IoU: $\%$) for U-Net, wide U-Net and our suggested architecture UNet++ with and without deep supervision (DS).}
\label{tab:main_results}
    \begin{tabular}{P{0.23\linewidth}P{0.12\linewidth}P{0.15\linewidth}P{0.15\linewidth}P{0.15\linewidth}P{0.15\linewidth}}
    \hline
    \multirow{2}*{Architecture} & \multirow{2}*{Params} & \multicolumn{4}{c}{Dataset} \\
    \cline{3-6}
     & & cell nuclei & colon polyp & liver & lung nodule \\
    \hline
    U-Net~\cite{ronneberger2015u}   & 7.76M & 90.77 & 30.08 & 76.62 &71.47 \\
    Wide U-Net & 9.13M  & 90.92 & 30.14 & 76.58 & 73.38\\
    UNet++ w/o DS & 9.04M &   \textbf{92.63}&	\textbf{33.45}&	79.70&	76.44\\
    UNet++ w/ DS & 9.04M&	92.52&	32.12&	\textbf{82.90}& \textbf{77.21}\\
    \hline
    \end{tabular}
\end{threeparttable}
\end{center}
\end{table}

\section{Conclusion}
\label{sec:conclusion}

To address the need for more accurate medical image segmentation, we proposed UNet++. The suggested architecture takes advantage of re-designed skip pathways and deep supervision. The re-designed skip pathways aim at reducing the semantic gap between the feature maps of the encoder and decoder sub-networks, resulting in a possibly simpler optimization problem for the optimizer to solve. Deep supervision also enables more accurate segmentation particularly for lesions that  appear at multiple scales such as polyps in colonoscopy videos. We evaluated UNet++ using four medical imaging datasets covering lung nodule segmentation, colon polyp segmentation, cell nuclei segmentation, and liver segmentation. Our experiments demonstrated that UNet++ with deep supervision achieved an average IoU gain of 3.9 and 3.4 points over U-Net and wide U-Net, respectively.


\subsubsection*{Acknowledgments}

This research has been supported partially by NIH under Award Number R01HL128785, by ASU and Mayo Clinic through a Seed Grant and an Innovation Grant. The content is solely the responsibility of the authors and does not necessarily represent the official views of NIH.

{\small
\bibliographystyle{ieee}
\bibliography{DLMIA,PAMI_JL}
}

\end{document}